\documentclass{article}

\usepackage[nonatbib, preprint]{neurips_2024}

\usepackage[utf8]{inputenc} 
\usepackage[T1]{fontenc}    
\usepackage{hyperref}       
\usepackage{url}            
\usepackage{booktabs}       
\usepackage{amsfonts}       
\usepackage{nicefrac}       
\usepackage{microtype}      
\usepackage{xcolor}         
\usepackage{graphicx}
\usepackage{amsmath}
\title{Scaling Law Hypothesis for Multimodal Model}

\author{%
  Bill (Qingyun) Sun\\
  PIN AI and G-Alpha\\
  Hillsborough CA\\
  \texttt{bill@pinai.io} \\
  \And
  Gavin (Zhen) Guo\thanks{Currently at Apple}\\
  MIT EECS\\
  Cambridge, MA\\
  \texttt{zguo0525@mit.edu} \\
  \And
  PIN AI Team\\
  \href{https://www.pinai.io/}{pinai.io}\\
  San Francisco, CA\\
  \texttt{www.pinai.io}
}

\begin{document}
\maketitle

\begin{abstract}
We propose a scaling law hypothesis for multimodal models processing text, audio, images, and video within a shared token and embedding space. Our framework predicts model performance based on modality-specific compression and tokenization efficiency, extending established scaling laws from text-based decoder models to mixed-modality systems. We explore whether leveraging more training data in multiple modalities can reduce the size of the multimodal model, enabling efficient deployment on resource-constrained devices.
\end{abstract}

\section*{Introduction}

Scaling laws in large language models (LLMs) have unveiled fundamental relationships between model performance, size, and the volume of training data~\cite{kaplan2020scalinglawsneurallanguage, hoffmann2022training, anil2023palm, rae2022scaling, muennighoff2023scaling, sardana2023chinchillaoptimal}. These laws serve as a guide for resource allocation in LLM development, helping to balance model size and data volume to optimize performance. The initial scaling laws proposed by OpenAI~\cite{kaplan2020scalinglawsneurallanguage} suggested that larger models are more sample-efficient, leading to the creation of massive models like GPT-3. However, subsequent research from DeepMind, notably the Chinchilla study~\cite{hoffmann2022training}, revealed that many large models were undertrained. Their findings indicated that smaller models trained on more data could outperform larger models when the compute budget is held constant.

Despite these insights, recent trends challenge the Chinchilla-optimal law. For instance, models like Llama 3 and 3.1 have been trained on significantly more tokens (up to 10 times more than Chinchilla's recommendations), yet still demonstrate outstanding performance~\cite{meta_llama_3}. This discrepancy has prompted researchers to reconsider the optimal allocation of compute resources in autoregressive pre-training~\cite{guo2024computeneed}.

\begin{figure}[h!]
\centering
\includegraphics[width=0.9\textwidth]{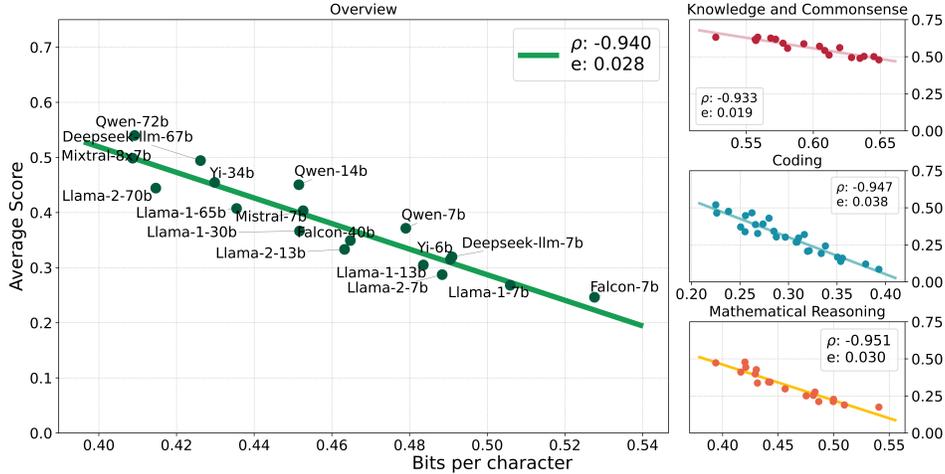}
\caption{Performance (measured by BPC) correlates with data compression ability across different text corpora.}
\label{fig1}
\end{figure}

Recent work suggests a unified scaling law, where model performance is driven primarily by total compute, regardless of how it is distributed between model size and dataset size~\cite{guo2024computeneed}. This approach introduces bits per character (BPC) as a performance metric that reflects the model's compression efficiency~\cite{huang2024compression}. BPC has been shown to correlate linearly with model performance across various modalities (Figure \ref{fig1}).

\begin{figure}[h!]
\centering
\includegraphics[width=0.85\textwidth]{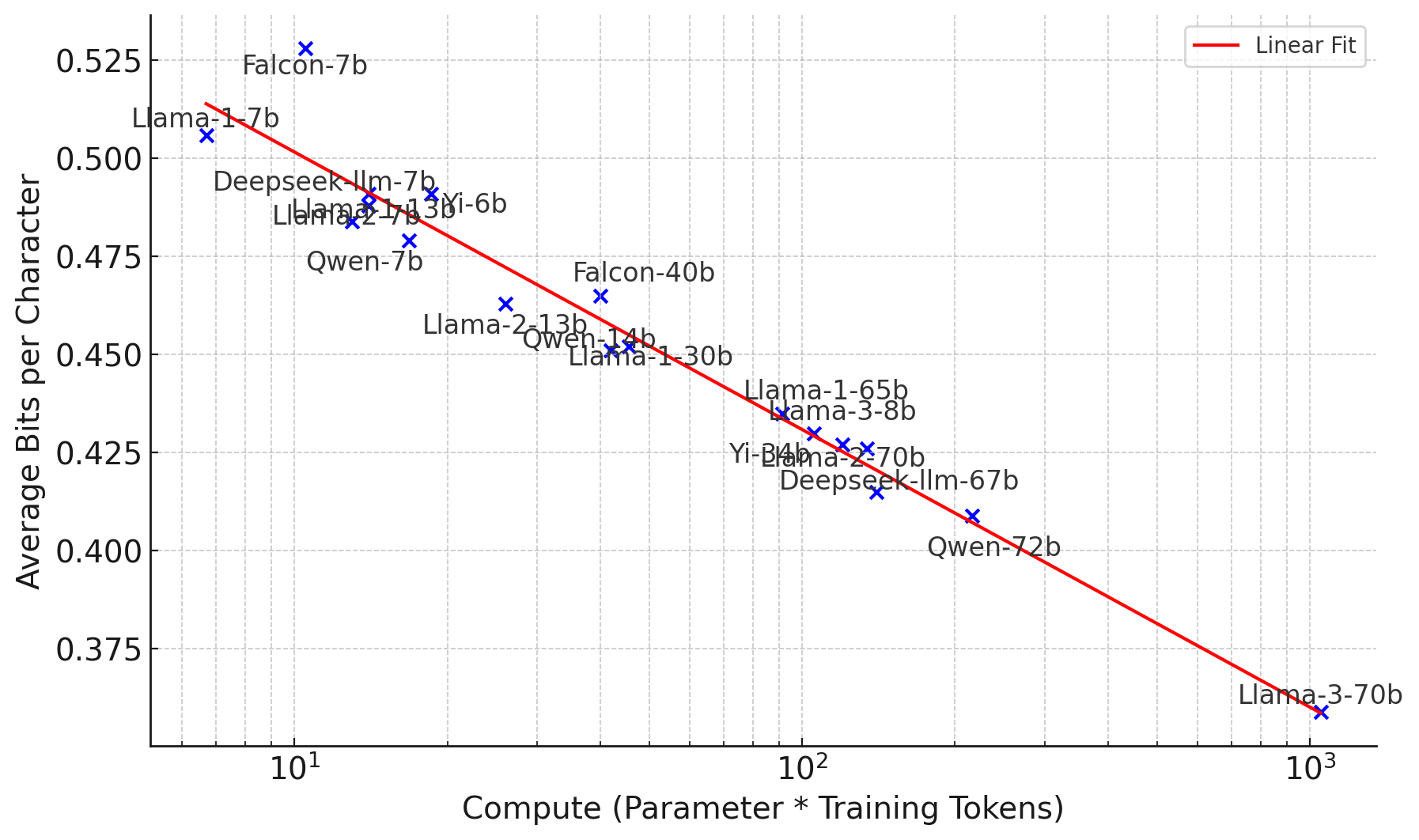}
\caption{Performance scales linearly with compute when measured by BPC~\cite{guo2024computeneed}.}
\label{fig2}
\end{figure}

This perspective reveals a linear relationship between BPC and the logarithm of compute used, which can be formalized as:

\begin{equation}
\text{BPC} \propto \log(N) + \log(P)
\end{equation}

where \( N \) is the number of training tokens, and \( P \) is the number of model parameters (Figure \ref{fig2}).

This unified scaling law suggests that smaller models trained on larger datasets may be prioritized for inference efficiency, especially in settings where resource constraints in inference are significant.

In multimodal systems, diverse types of data, such as text, audio, images and video, are processed through specialized tokenization techniques, each with varying levels of compression efficiency. Text is tokenized using methods like Byte Pair Encoding (BPE)~\cite{sennrich2016neuralmachinetranslationrare}, which offers relatively stable compression efficiency. Audio data is typically handled through spectrogram-based methods~\cite{langman2024spectralcodecsspectrogrambasedaudio}, while images and video, which are more redundant and higher-dimensional, require convolutional neural networks (CNNs), vision transformers~\cite{dosovitskiy2021imageworth16x16words, han2022survey}, and spatio-temporal transformers~\cite{arnab2021vivit, liu2022video, selva2023video} for tokenization. These tokenized inputs are unified in a shared token and embedding space, enabling multimodal models to process them together.

The motivation for our work stems from the need to extend scaling laws, previously established for text-based models, to multimodal models. Current scaling laws have guided the development of large models in natural language processing, demonstrating how model performance scales with increased data and model size. However, these insights have not been fully explored in the context of multimodal models, where different modalities have varying tokenization efficiencies and data complexities.

Our goal is to investigate whether increasing the amount of training data across multiple modalities can compensate for reducing the size of the multimodal model. If this hypothesis holds, it would allow smaller models to perform comparably to larger models, particularly in resource-constrained environments such as mobile or edge devices. By optimizing the trade-off between model size and data efficiency, we aim to enable multimodal models that are more computationally feasible for on-device applications without sacrificing performance.

\section*{Modality-Specific Compression and Tokenization Efficiency}  
The compression and tokenization efficiency of each modality is a key determinant of overall model performance. For each modality \(i\), the relationship between the raw input size \(T_i\), the compression factor \(C_i\), and the number of tokens generated \(N_i\) can be expressed as:

\begin{equation}
\log T_i = \log C_i + \log N_i,
\end{equation}

where \(C_i\) represents the efficiency with which continuous data is compressed into tokens, and \(N_i\) is the number of tokens produced after compression.

In text-based models, tokenization efficiency is relatively stable due to the consistency of algorithms like Byte Pair Encoding (BPE). This consistency allows scaling laws to focus primarily on the number of text tokens \(N_{\text{text}}\) and model size \(P\) as performance predictors. However, in multimodal models, tokenization efficiency varies significantly between modalities due to differences in the complexity and redundancy of raw input data.

For example, visual data such as images tends to have a high level of redundancy, allowing for significant compression. Nevertheless, tokenizing image data through CNNs or vision transformers results in a larger number of tokens compared to text, primarily due to the higher dimensionality of images. Video data, which combines both spatial and temporal dimensions, is even more complex and generates a greater number of tokens, further increasing the compute required.

This variability in compression efficiency across modalities underscores the need for a scaling law that accounts for the specific characteristics of each data type, ensuring that model performance is accurately predicted for mixed-modality systems.

\section*{Predicting Multimodal Model Performance}  
For text-only models, performance typically scales according to:

\begin{equation}
    \text{performance} \propto \log(N_{\text{text}}) + \log(P),
\end{equation}

where \(N_{\text{text}}\) is the number of text tokens, and \(P\) is the number of model parameters~\cite{guo2024computeneed}.

When this scaling law is extended to multimodal models, it must account for the differing compression efficiencies across modalities. The performance of multimodal models is determined by the total amount of raw data represented in the shared token space, adjusted by the compression efficiency of each modality. The performance can be predicted by the following equation:

\begin{equation}
\text{multimodal performance} \propto \sum_i \log \left(  \frac{T_i}{C_i} \right) + \log P,
\end{equation}

where \(T_i\) represents the raw data size for each modality, \(C_i\) is the compression efficiency for that modality, and \(P\) is the number of model parameters.

This equation highlights that while the total raw data processed by the model contributes to overall performance, the efficiency of tokenization for each modality (i.e., how effectively each modality compresses its raw data into tokens) plays a significant role in the compute required. A modality with lower compression efficiency, such as video, will require substantially more compute to reach the same performance level as text, which generally has higher compression efficiency.

\section*{Conclusion}  
We propose a scaling law hypothesis that extends established text-based scaling laws to multimodal models, emphasizing the crucial role of modality-specific compression and tokenization efficiency. The performance of multimodal models depends not only on the total amount of raw data and model size but also on how efficiently each modality compresses its data into tokens. This relationship directly affects the computational resources required to train the model.

Our hypothesis explores the potential to leverage larger amounts of training data across multiple modalities to reduce model size without sacrificing performance. This could enable more efficient multimodal models, especially in resource-constrained environments like mobile devices or edge computing. By optimizing the trade-off between data volume and model size, we aim to make multimodal models more suitable for on-device deployment.

Future work should focus on refining the quantification of compression factors for each modality, allowing for more accurate performance predictions and guiding the development of optimized multimodal architectures for a wide range of tasks and data types.

\section*{Limitations}
The proposed scaling law assumes that the multimodal model is trained from scratch. It may not directly apply to models that utilize cross-modal connectors, such as LLaVA~\cite{liu2023llava, liu2023improvedllava} and VILA~\cite{lin2023vila, xue2024longvilascalinglongcontextvisual}, which align pre-trained vision and language models. These approaches leverage pre-trained components, which could affect the scaling dynamics and performance predictions outlined in this work.

\bibliography{ref}
\bibliographystyle{unsrt}

\end{document}